\documentclass[journal,twoside,web]{ieeecolor}
\usepackage{generic}
\usepackage{cite}
\usepackage{amsmath,amssymb,amsfonts}
\usepackage{graphicx}
\usepackage{algorithm}
\usepackage{hyperref}
\usepackage{cite}
\usepackage{amsmath,amssymb,amsfonts}
\usepackage{graphicx}
\usepackage{textcomp}
\usepackage{xcolor}
\usepackage{booktabs}
\usepackage{algpseudocode}
\usepackage{booktabs}
\usepackage{multirow}
\usepackage{tabularx}
\usepackage{array}
\usepackage{makecell}
\usepackage{caption}
\usepackage{hyperref}
\newcolumntype{Y}{>{\raggedright\arraybackslash}X} 
\usepackage{makecell} 
\usepackage{rotating}
\hypersetup{hidelinks=true}
\usepackage{textcomp}
\def\BibTeX{{\rm B\kern-.05em{\sc i\kern-.025em b}\kern-.08em
    T\kern-.1667em\lower.7ex\hbox{E}\kern-.125emX}}
\begin{document}
\title{MMIR-TCM: Memory-Integrated Multimodal Inference and Retrieval for TCM Clinical Decision Support}
\author{Lihui Luo, Joongwon Chae, Ziyan Chen, Yang Liu, Siyi Cheng, Weihan Gao, Zelin Zeng, Xiaoming Yin, Samaneh Beheshti Kashi, Dongmei Yu, Lian Zhang, Jing Sui, Zeming Liang, Jiansong Ji, Peter E. Lobie, and Peiwu Qin \thanks{Lihui Luo, Joongwon Chae, Ziyan Chen, Yang Liu, Siyi Cheng, Weihan Gao and Peter E. Lobie are with the Institute of Biopharmaceutics and Health Engineering, Tsinghua Shenzhen International Graduate School, Shenzhen, 518000, China, also with the Chinese Medicine Guangdong Laboratory, Zhuhai, 519000, China (e-mail: luolh23@mails.tsinghua.edu.cn; chi-zy24@mails.tsinghua.edu.cn; chenziya23@mails.tsinghua.edu.cn; lyang22@mails.tsinghua.edu.cn; chengsiy25@mails.tsinghua.edu.cn; gwh25@mails.tsinghua.edu.cn; pelobie@sz.tsinghua.edu.cn).} 
\thanks{Jing Sui is with the Beijing Normal University, Beijing, 100000, China (e-mail:Xi0218@hotmail.com).}
\thanks{Zeming Liang and Peiwu Qin are with the Chinese Medicine Guangdong Laboratory, Zhuhai, 519000, China (e-mail: 305823980@qq.com, pwqin1979@gmail.com).} 
\thanks{Lian Zhang is with the First Hospital of Hebei Medical University, Shijiazhuang, 050000, China (e-mail: pwqin@sz.tsinghua.edu.cn).} 
\thanks{Samaneh Beheshti Kashi, Dongmei Yu and Jiansong Ji are with the Lishui Hospital of Zhejiang University, Lishui, 323000, China (e-mail:lolisky1@163.com, qaydm1979@gmail.com, llzzmm0218@163.com).}
\thanks{Zelin Zeng and Xiaoming Yin are with XiaoMing TCM Hospital, Shenzhen, 518000 China. (e-mail: 13074519056@163.com; 15864642233@163.com).} \thanks{The code is available at \href{https://github.com/jw-chae/MMIR-TCM}{https://github.com/jw-chae/MMIR-TCM}.} \thanks{Lihui Luo and Joongwon Chae contributed equally to this work.} \thanks{Corresponding author: Peiwu Qin.}}

\maketitle

\begin{abstract}
Traditional Chinese Medicine (TCM) diagnosis, particularly through tongue inspection, faces persistent challenges in subjectivity and reproducibility. The application of multimodal artificial intelligence to TCM clinical tasks, such as syndrome differentiation and prescription generation, is significantly hampered by the semantic gap between visual tongue features and textual reasoning, as well as the lack of large-scale, standardized datasets. To address these challenges, we introduce MMIR-TCM, a novel framework that emulates the diagnostic process of TCM experts by integrating multimodal large language model(MLLM) with memory-augmented segmentation and retrieval-augmented generation (RAG). Employing a three-stage architecture, MMIR-TCM integrates a training-free Memory-SAM module for robust tongue extraction, a fine-tuned Qwen3-VL model for structured tongue diagnosis generation, and a Qwen3-based RAG component for evidence-grounded clinical decision support generation. The framework was developed and validated using MedTCM, a new large-scale multimodal dataset that we introduce specifically for advanced TCM research. To properly evaluate our framework's clinical accuracy, which existing metrics fail to capture, we also developed TDEU, a domain-specific evaluation metric incorporating semantic understanding and diagnostic importance. Our comprehensive experiments demonstrate that MMIR-TCM significantly outperforms leading models, including GPT-4o and Gemini 2.5 Flash.

\end{abstract}

\begin{IEEEkeywords}
Artificial Intelligence, Large Language Models, Traditional Chinese Medicine, Tongue Image Analysis, Clinical Decision Support Systems.
\end{IEEEkeywords}

\section{Introduction}
\IEEEPARstart{M}{ultimodal} Artificial Intelligence  (AI) have recently achieved radiologist-level performance on several Western medical imaging tasks, such as chest X ray interpretation and abnormality detection~\cite{ryu2025vision}. However, Traditional Chinese Medicine (TCM), a diagnostic paradigm built on holistic pattern recognition and centuries of clinical practice, remains relatively underexplored by modern AI. TCM diagnosis tightly couples visual observations (tongue, facial complexion), auditory, tactile cues (auscultation and pulse), and textual knowledge (syndrome typologies, herbal properties, and classical case precedents). This inherent multimodality and the need for evidence grounded clinical decisions make TCM a natural, yet challenging, target for multimodal model integration. 

Within the Four Diagnostic Methods of TCM (inspection, auscultation and olfaction, inquiry, and palpation), tongue inspection holds a special diagnostic role. Morphological features of the tongue body (e.g., color, shape, moisture), and features of the coating (e.g., color, thickness, texture, distribution) are systematically associated with core pathophysiological patterns \cite{wu2020tongue}. For example, a pale tongue with a thin white coating may indicate qi deficiency and internal cold, suggesting the use of tonifying formulas such as \emph{Si Junzi Tang}, whereas a red tongue with a yellow greasy coating implies damp–heat, prompting clearing and draining strategies like \emph{Longdan Xiegan Tang}. Clinicians synthesize tongue visual cues, patient history, and canonical case precedents to choose personalized herbal prescriptions \cite{segawa2023objective}.

Despite its clinical significance, tongue diagnosis suffers from several challenges that limit its reproducibility and computational integration. Practitioner subjectivity introduces considerable inter- and intra-rater variability \cite{arji2019systematic}. In addition, variations in image acquisition conditions, such as illumination (e.g., changes in color temperature), background clutter (e.g., clothing, facial features), and framing inconsistencies (e.g., tongue positioning, camera angle) can substantially degrade the reliability of automated analysis pipelines. Moreover, existing approaches to tongue image analysis~\cite{liu2023survey} typically address segmentation or attribute classification in isolation, without bridging the gap to clinical decision-making or grounding outputs in verifiable knowledge sources.

The emergence of multimodal large language models (MLLMs) offers a promising new approach. Leveraging strong image understanding capabilities, MLLMs have shown competitive performance in a variety of medical imaging tasks, including disease classification and visual report generation. For instance, Zhao et al. \cite{zhao2023pregenerator} fine tuned MLLMs by integrating visual data with structured medical knowledge, enabling clinical reasoning within diagnostic workflows. However, the application of large language models (LLMs) to TCM tasks is constrained by the scarcity of large scale, publicly available multimodal datasets. Existing resources tend to be fragmented, providing isolated elements such as herbal prescriptions or tongue images rather than complete clinical records that pair images with diagnoses and treatment strategies \cite{zhang2025large}.

Motivated by these considerations, we propose MMIR-TCM, an end-to-end multimodal framework that emulates the staged reasoning workflow of experienced TCM practitioners: memory-integrated tongue region extraction, structured tongue diagnosis generation, and retrieval-augmented-generation (RAG) \cite{lu2024clinicalrag} for prescription generation that grounds clinical decisions in precedent and evidence. To support rigorous evaluation and reproducibility, we assemble MedTCM, a multi center multimodal corpus reflecting realistic clinical diversity, and introduce TDEU, a domain aware metric designed to evaluate tongue diagnostic outputs with attention to both semantic fidelity and clinical importance. Together, these components enable a system that not only predicts tongue diagnosis accurately but also links those predictions to verifiable and evidence-anchored clinical recommendations. By grounding generation in retrieved clinical precedents and documented formula rationales, the system improves factuality, interpretability, and clinician auditability, which are essential for safe clinical deployment.

In summary, the main contributions of this work are as follows:
\begin{enumerate}
\item We propose MMIR-TCM, a memory-integrated multimodal pipeline for TCM clinical decision support that integrates training free tongue extraction, attribute-level tongue diagnosis generation, and retrieval-augmented prescription generation to produce evidence-grounded clinical decision support.
\item We construct MedTCM, a large scale multimodal TCM dataset with tongue images, diagnostic reports, and clinical prescriptions collected from multiple hospitals to ensure diversity and clinical relevance. The dataset will be made publicly available and maintained.
\item We develop TDEU, a domain aware evaluation metric for tongue diagnosis that incorporates semantic similarity and clinical importance, addressing the limitations of traditional text matching metrics.
\end{enumerate}

\begin{figure*}[t]
\centering
\includegraphics[width=0.95\textwidth]{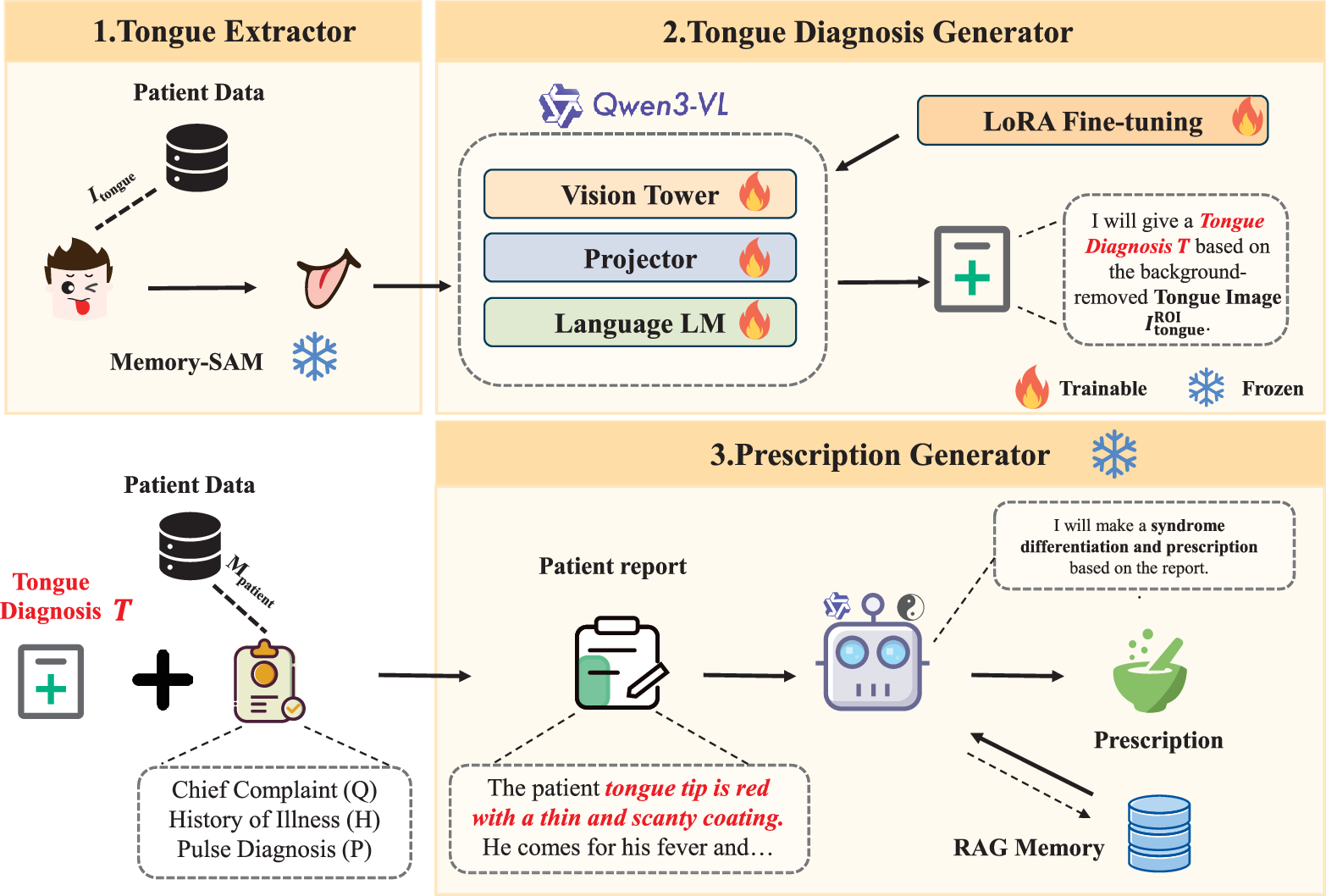}
\caption{Overall architecture of MMIR-TCM. A Memory-SAM–based Tongue Extractor precedes a Qwen3-VL tongue diagnosis generator, followed by a Qwen3-based RAG prescription generator, emulating expert TCM clinical workflow.}
\label{fig:overall}
\end{figure*}
\section{Related Work}

\subsection{TCM-Specific Large Language Models}

Recent efforts have adapted large language models (LLMs) to TCM through domain-specific pretraining and alignment. 
TCMChat~\cite{dai2024tcmchat} demonstrated that continued pretraining on 
TCM corpora followed by supervised fine-tuning improves knowledge recall 
and question-answering over general-purpose LLMs. Complementary benchmark 
efforts have emerged to standardize evaluation: TCMBench~\cite{yue2024tcmbench} 
curated 5,473 licensing exam questions with the TCMScore metric to assess 
TCM semantic consistency beyond surface accuracy, revealing that current 
models still have considerable room for improvement. TCMD~\cite{yu2024tcmd} 
provided large-scale annotated QA with robustness analysis, exposing 
inconsistencies under random perturbations. BianCang~\cite{wei2025biancang} 
implemented a two-stage learning pipeline (knowledge injection followed by 
targeted alignment) using the Chinese Pharmacopoeia and hospital records, 
demonstrating superior performance in syndrome differentiation tasks. 
These efforts underscore the value of TCM-specific alignment, though 
excessive fine-tuning may degrade general language capabilities~\cite{yue2024tcmbench}.

For clinical prescription recommendation, recent work has moved beyond 
pure LLM generation toward knowledge-augmented approaches. 
TCM-FTP~\cite{zhou2024tcm} fine-tuned LLMs for herbal prescription 
prediction and introduced normalized mean squared error (NMSE) metrics 
for dosage prediction, directly addressing the quantification challenge 
in formula generation. TCM-KLLaMA~\cite{zhuang2025tcm} combined knowledge 
graphs encoding herb-symptom-contraindication relationships with LLM 
generation, using structured knowledge constraints to suppress interaction 
risks and improve groundedness.

\subsection{Multimodal Tongue Diagnosis}

Tongue diagnosis occupies a central role in TCM's Four Examinations 
(inspection, auscultation, inquiry, palpation), as properties 
of the tongue body (color, shape) and coating (color, thickness, texture, 
distribution) directly inform syndrome differentiation. Early computational 
approaches focused on segmentation and attribute classification using 
traditional computer vision or deep learning~\cite{liu2023survey}, but 
recent work has shifted toward multimodal fusion that jointly reasons 
over images and textual clinical information. TongueNet~\cite{zhou2019tonguenet} 
introduced image-text fusion with consistency and complementarity 
constraints in the representation space, enabling simultaneous multi-label 
prediction of pathological attributes and anatomical locations. 
Cross-modal attention architectures~\cite{gan2025cross} have 
extended this to organ-level pathology classification, bridging the gap 
between visual features and high-level clinical reasoning. However, 
robustness to acquisition variations—illumination shifts, color bias, 
and partial occlusion—remains underexplored in these frameworks.

Dataset standardization has emerged as a critical enabler for reproducible 
evaluation. TCM-Tongue~\cite{jin2025tcm} provides 6,719 high-quality 
images with 20 pathological annotations under controlled acquisition 
conditions, while TCMEval-SDT~\cite{wang2025tcmeval} offers a benchmark 
for syndrome diagnosis capability. These resources facilitate systematic 
comparison of multimodal models, though comprehensive robustness analysis 
across illumination, compression, and occlusion stressors is still limited.

\paragraph{Impact of Segmentation on Downstream Diagnosis.}
Accurate tongue--background separation is widely reported to improve downstream performance in classical vision pipelines.
For example, Xian et al.~\cite{xian2022automatic} showed that using segmentation as an auxiliary loss can enhance both quality assessment and subsequent diagnostic stages. 
For fine-grained structures, segment-based approaches~\cite{yan2023tongue} enable stable crack detection, and improved TransUNet variants~\cite{wu2024novel} achieve precise coating segmentation that stabilizes feature extraction.
Moreover, integrating quality control with segmentation can filter low-quality shots prior to analysis, thereby reducing variance and improving reproducibility.
Recent zero-shot methods such as TongueSAM~\cite{cao2023tonguesam} leverage Segment Anything to generalize across diverse acquisition conditions without retraining.

\subsection{Retrieval-Augmented Generation for Clinical Decision Support}

Retrieval-augmented generation (RAG) has become the standard approach 
for grounding medical LLM outputs in external knowledge sources. 
General RAG frameworks~\cite{gao2023retrieval} retrieve relevant documents 
to augment generation context, improving factual accuracy and reducing 
hallucination. In medical domains, specialized architectures have 
emerged: ClinicalRAG~\cite{lu2024clinicalrag} implements multi-agent 
workflows that dynamically integrate structured codes and unstructured 
clinical notes, while LINS~\cite{wang2025lins} enhances evidence-based 
medicine with citation consistency metrics. GraphRAG~\cite{he2025opentcm} 
extends retrieval to graph-structured knowledge, enabling multi-hop 
reasoning and query-focused summarization over large private corpora. 
Recent evaluations~\cite{gaber2025evaluating} confirm that RAG workflows 
outperform pure LLMs in clinical appropriateness and urgency assessment 
tasks.

In TCM, knowledge sources bifurcate into \textbf{empirical case 
knowledge} (modern EMRs) and \textbf{normative principle knowledge} 
(classical texts such as \emph{Huangdi Neijing}, \emph{Shanghan Lun}, 
\emph{Bencao Gangmu})~\cite{unschuld2011huang}. These sources exhibit 
representational mismatches—modern vernacular versus classical Chinese, 
symptom-syndrome-formula taxonomies—that complicate unified indexing. 
PreGenerator~\cite{zhao2023pregenerator} addressed this through a 
retrieval→generation hybrid that first recalls similar prescriptions 
and cases before combining them with generation rules, achieving 
consistent improvements over pure neural models. OpenTCM~\cite{he2025opentcm} 
constructed a TCM knowledge graph from 68 classical obstetrics texts 
(3.73M characters) with 48k entities and 152k relations, integrating 
GraphRAG for symptom-based retrieval and demonstrating expert-evaluated 
superiority in diagnostic QA. These approaches highlight the 
complementarity of case-based specificity and principle-based normativity, 
motivating our hybrid indexing strategy. Furthermore, integration with 
clinical practice guidelines~\cite{oniani2024enhancing} enables injection of 
normative constraints (contraindications, interaction rules) into the 
reasoning chain, which is critical for safety-critical prescription 
generation in TCM.

\subsection{Evaluation, Robustness, and Safety in Medical AI}

Traditional metrics such as BLEU and ROUGE capture surface-level 
similarity but fail to reflect clinical semantic equivalence or groundedness. 
TCM-specific benchmarks have introduced domain-aware evaluation: TCMBench's TCMScore~\cite{yue2024tcmbench} 
incorporates semantic consistency, while MTCMB~\cite{kong2025mtcmb} 
decomposes evaluation across 12 subtasks including knowledge QA, 
clinical reasoning, prescription generation, and safety compliance, 
revealing that current LLMs perform adequately on knowledge recall 
but struggle with clinical reasoning and safety adherence. Multimodal 
evaluation has expanded to include tongue and herb images, audio, and 
video through TCM-Ladder~\cite{xie2025tcm}, measuring cross-modal 
grounding capabilities beyond text accuracy.

For RAG systems, evaluation must extend beyond answer quality to 
retrieval and augmentation stages. RAG frameworks such as RAGAS~\cite{es2024ragas} propose metrics for 
faithfulness (evidence-output alignment), answer relevance, and context 
relevance, with faithfulness showing strong correlation with human 
judgments. GroUSE~\cite{muller2024grouse} benchmarks 
judge LLMs themselves, exposing failure modes such as partial citation 
and context overfitting. Medical applications demand additional rigor: 
MedCite~\cite{wang2025medcite} targets "verifiable text" in medical QA 
through multi-pass retrieval and citation, elevating citation 
consistency to a first-class output criterion.

Robustness, explainability, and safety form the operational pillars 
for clinical deployment. Robustness encompasses resilience to input 
perturbations (noise, illumination shift, occlusion), distribution 
shift (out-of-distribution detection~\cite{hong2024out}), and 
reasoning chain vulnerabilities~\cite{balendran2025scoping}. 
Uncertainty quantification techniques—temperature scaling, deep 
ensembles, conformal prediction~\cite{lambert2024trustworthy,fayyad2024empirical}—enable 
selective prediction with rejection thresholds for high-risk cases. 
Explainability in medical contexts requires \textbf{attribute-level 
grounding}: attention maps, segmentation masks, and counterfactual 
interventions~\cite{singla2023explaining} to validate feature-decision 
pathways. Safety enforcement in prescription generation necessitates 
multi-layered defenses: (1) pretraining safety alignment with 
contraindication rules, (2) constrained decoding with normative 
scoring~\cite{fu2024constrained}, (3) post-hoc validators using 
rule-based and knowledge-graph checking, 
and (4) human-in-the-loop gates for high-risk decisions~\cite{asgari2025framework}. 
Systematic reviews emphasize that accuracy alone is insufficient; 
trustworthy medical AI must co-optimize reliability, interpretability, 
and accountability~\cite{tam2024framework}.

\section{Methods}
We present MMIR-TCM, an end-to-end multimodal framework that mirrors expert TCM workflows by standardizing visual inputs, generating structured tongue descriptions, and producing evidence-grounded prescriptions via retrieval-augmented reasoning. As illustrated in Fig.~\ref{fig:overall}, the system processes tongue images and clinical metadata jointly to yield accurate and interpretable outputs. Given a patient's tongue image $I_{\text{tongue}}$ and clinical metadata $M_{\text{patient}} = (Q, H, P)$, where $Q$, $H$, and $P$ denote chief complaint, history of present illness, and pulse diagnosis, respectively, the pipeline consists of three tightly coupled stages:

\begin{enumerate}
    \item \textbf{Tongue Extractor.} A training-free Memory-SAM segmenter retrieves visual exemplars and converts them into structured foreground/background point prompts for SAM2~\cite{ravi2024sam}, yielding a precise tongue mask and a background-removed region-of-interest (ROI) $I_{\text{tongue}}^{\mathrm{ROI}}$.
    \item \textbf{Tongue Diagnosis Generator.} A fine-tuned Qwen3-VL model analyzes $I_{\text{tongue}}^{\mathrm{ROI}}$ and outputs a one-sentence, attribute-level report $D$ that consistently enumerates tongue body color/shape and coating color/thickness/texture with anatomical locations.
    \item \textbf{Prescription Generator.} A Qwen3-based generator fuses $D$ with $M_{\text{patient}}$, retrieves similar cases from a de-identified EMR knowledge base via vector search, and synthesizes syndrome differentiation and an herbal prescription with concise evidence-based reasoning.
\end{enumerate}

This modular design enables independent optimization of each stage while preserving end-to-end traceability from raw images to clinically actionable, evidence-grounded outputs.

\begin{figure}[h]
\includegraphics[width=\columnwidth]{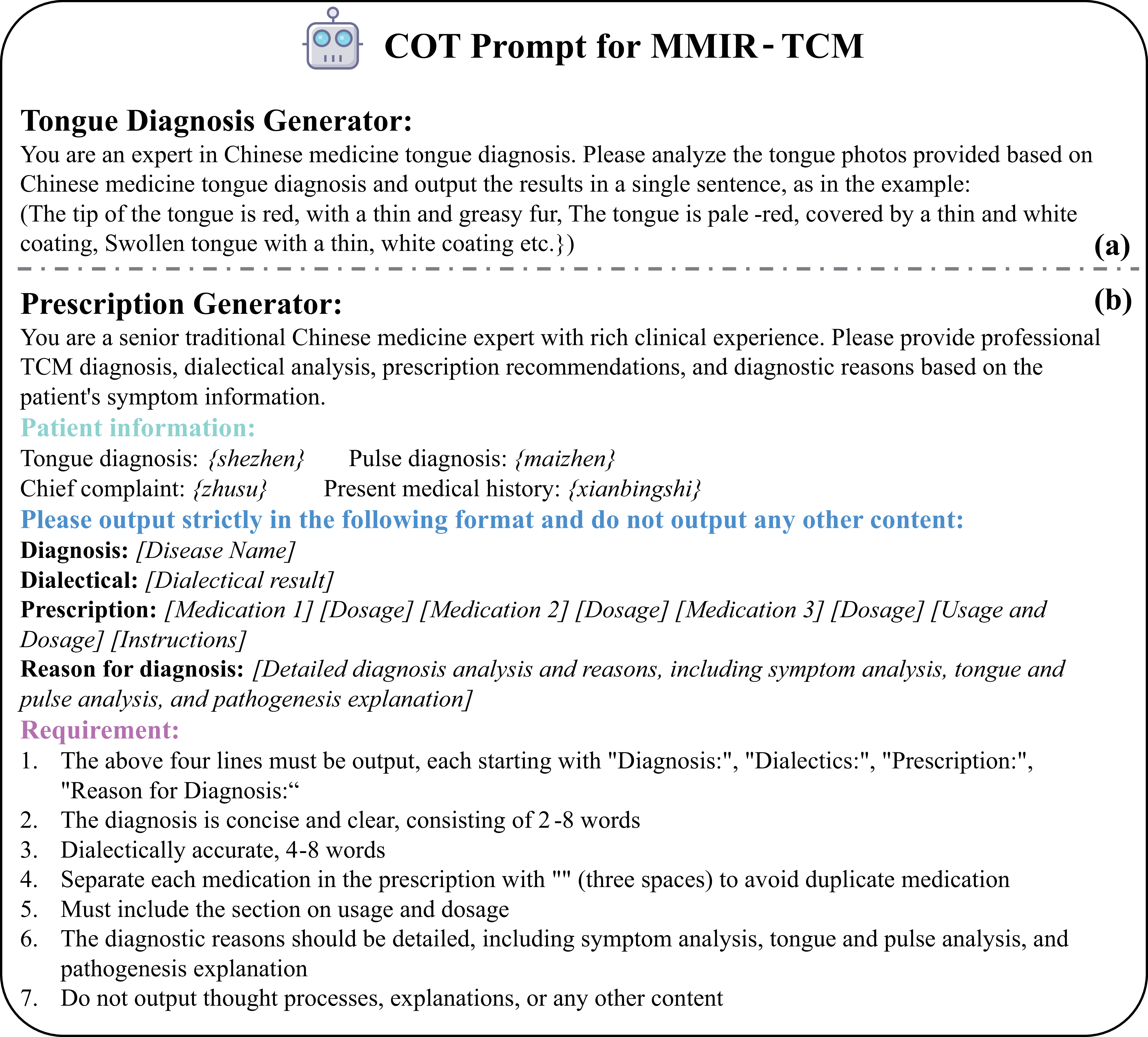}
\caption{The detailed prompt template used in MMIR-TCM.}
\label{prompt}
\end{figure}
\subsection{Training-Free Tongue Extractor via Memory-SAM}

Supervised segmentation models require extensive pixel-level annotations 
and exhibit poor generalization to out-of-distribution acquisition 
conditions—varying lighting, background clutter, camera specifications, 
and patient positioning. Bounding-box-based pipelines that feed detected 
regions to SAM suffer from detector failures under these conditions, 
often leaking into surrounding facial regions. To address these limitations 
without incurring annotation costs or retraining overhead, we used 
Memory-SAM\cite{chae2025memory}, a training-free architecture that converts visual exemplar 
retrieval into explicit point prompts for SAM2.


\paragraph{Dense Feature Extraction.}
Given a tongue image $I_{\text{tongue}}$, we extract patch embeddings using a pretrained 
DINOv3 (ViT-L/16) encoder~\cite{simeoni2025dinov3}. Each patch embedding 
is $\ell_2$-normalized to form a dense feature map $E_q \in \mathbb{R}^{H 
\times W \times D}$, where $H, W$ are spatial dimensions and $D=1024$ is 
the embedding dimension. We also compute a global image descriptor 
$\bar{e}_q \in \mathbb{R}^D$ by average-pooling patch embeddings.

The memory bank $\mathcal{M}$ stores $N$ reference exemplars, where each 
entry $m = (I_m, M_m, E_m, \bar{e}_m)$ consists of the reference image, 
ground-truth binary mask, patch embeddings, and global descriptor.

\paragraph{Memory Retrieval.}
We retrieve the most similar exemplar via cosine similarity over global 
descriptors:
\begin{equation}
m^* = \arg\max_{m \in \mathcal{M}} \text{sim}(\bar{e}_q, \bar{e}_m),
\end{equation}
where $\text{sim}(\cdot, \cdot)$ denotes cosine similarity. This top-1 
retrieval is implemented using FAISS~\cite{douze2025faiss} with a nearest-neighbor search.

\paragraph{Mask-Constrained Dense Matching.}
We resize $M_{m^*}$ to the feature grid resolution and define foreground 
and background index sets $\mathcal{J}_{\text{fg}}$ and $\mathcal{J}_{\text{bg}}$ 
based on mask values. For each query patch $i$, we compute pairwise 
cosine similarities with all reference patches:
\begin{equation}
S_{ij} = \frac{E_q^{(i)} \cdot E_{m^*}^{(j)}}{\|E_q^{(i)}\| \|E_{m^*}^{(j)}\|}.
\end{equation}

We then identify the best-matching foreground and background patches:
\begin{align}
j_{\text{fg}}^*(i) &= \arg\max_{j \in \mathcal{J}_{\text{fg}}} S_{ij}, \\
j_{\text{bg}}^*(i) &= \arg\max_{j \in \mathcal{J}_{\text{bg}}} S_{ij}.
\end{align}

\subsection{Attribute-Level Tongue Diagnosis Generator with Qwen3-VL}

Free-text tongue descriptions from practitioners vary widely in granularity, 
terminology, and structure, creating a semantic gap that obstructs downstream 
retrieval and reasoning. Inconsistent linguistic representations—ranging from 
telegraphic notes ("red tip, greasy coat") to verbose narratives—reduce the 
effectiveness of similarity-based retrieval and complicate evidence integration. 
To standardize the perception-reasoning interface, we fine-tune Qwen3-VL to 
generate consistent, attribute-level one-sentence tongue reports that 
enumerate key diagnostic features in a fixed format.

\subsubsection{Model and Training Objective}

We employ Qwen3-VL-30B~\cite{yang2025qwen3}, a vision-language model with 
a vision encoder (SigLIP), projection layer, and autoregressive language 
decoder. Given a tongue image $I_{\text{tongue}}$ and target report $T$, 
we minimize the cross-entropy loss:
\begin{equation}
\mathcal{L}(\theta) = -\sum_{(I,T) \in \mathcal{D}} \log P_\theta(T \mid I),
\end{equation}
where $\mathcal{D}$ is our training dataset of 2,805 image-report pairs 
from the MedTCM corpus, split 90/10 for training and validation.

To reduce computational overhead, we apply Low-Rank Adaptation 
(LoRA)~\cite{hu2022lora} to the vision projection layer and language decoder's 
query, key, value, and output projection matrices, as well as feed-forward 
network layers. We use LoRA rank $r \in \{64\}$ with 
$\alpha = 2r$. Training employs AdamW optimizer with learning rate 
$5 \times 10^{-5}$, weight decay 0.01, and fp16 mixed precision over 
3–20 epochs.

\subsubsection{Structured Output Format}

We enforce a standardized output schema through system prompts (See Fig. \ref{prompt} (a)).

Example output: \textit{"Tongue body pale red, shape normal, coating white, 
thin, slightly greasy, distributed evenly with mild redness at tip."}

This format ensures that every report covers the same diagnostic dimensions, 
enabling reliable parsing and structured input for downstream RAG retrieval.

\subsubsection{Implementation Details}

We use Qwen3-VL-30B with LoRA applied to both vision and language components 
(\textbf{rank 64}, 10 epochs, \textbf{dropout = 0}); the optimizer uses a learning rate 
of \textbf{$5 \times 10^{-5}$}. Input images are resized to $768 \times 768$ pixels. 
During inference, we use nucleus sampling with $p=0.9$ and temperature 
$T=0.7$ to balance consistency and natural language fluency.

\begin{figure*}[h]
\centering
\includegraphics[width=0.95\textwidth]{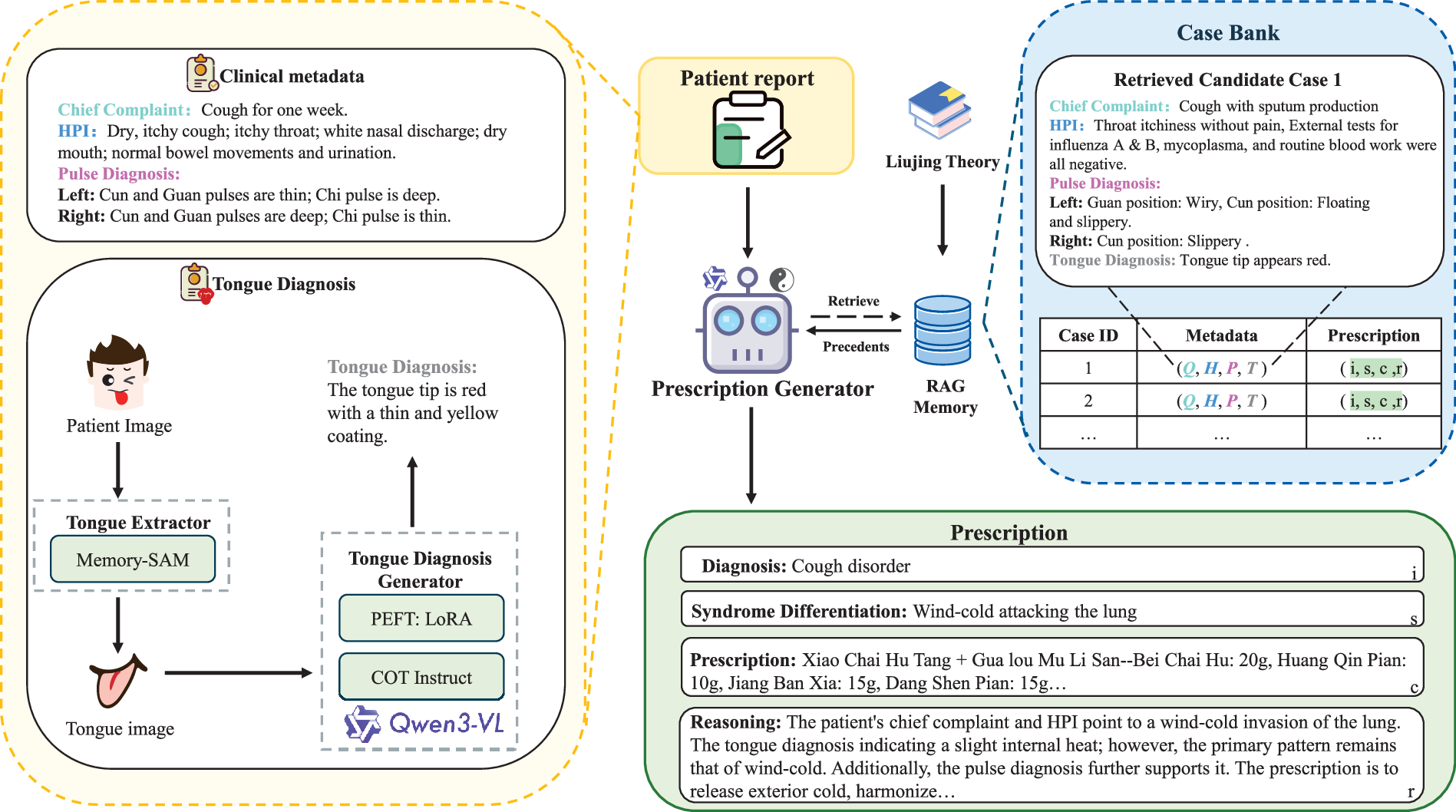}
\caption{RAG-Enhanced Prescription Generator. Retrieved similar cases and Liujing theory are fused with current patient metadata and tongue diagnosis to produce evidence-based clinical decision support.}
\label{rag}
\end{figure*}

\subsection{RAG-Enhanced Prescription Generator with Qwen3}
The diagnostic reasoning module is built upon a RAG framework grounded in Qwen3, designed to emulate the comprehensive and evidence-driven reasoning process of experienced TCM physicians. As shown in Fig.~\ref{rag}, this module integrates multimodal diagnostic information, including the generated tongue diagnosis $T_{\text{tongue}}$ and the patient's structured metadata $M_{\text{patient}}$, to retrieve clinically similar cases and synthesize context-aware prescriptions through LLM inference.

\subsubsection{Knowledge Base Construction and Indexing}
The foundation of our RAG system is a comprehensive Memory Base, consisting of two key components: the LiuJing Theory and the Clinical Case Bank. The case bank is built from 124,593 anonymized patient records collected from four major TCM hospitals. Each record contains complete clinical information—including chief complaints, medical history, tongue and pulse diagnoses, syndrome differentiation, and herbal prescriptions—providing rich contextual knowledge for retrieval. In parallel, the LiuJing Theory Memory incorporates canonical TCM theoretical texts, particularly those based on the LiuJing (Six-Meridian) differentiation, enabling the model to perform higher-level reasoning grounded in traditional diagnostic principles.

To enable efficient similarity-based retrieval, we employ Langchain, a data framework for LLM applications, combined with FAISS for vector indexing. The indexing process involves:
 
\begin{enumerate}
    \item \textbf{Document Processing}: Each record in memory base is structured as a document containing multiple fields (chief complaint, history, pulse, tongue, syndrome, prescription).
    
    \item \textbf{Semantic Embedding}: Documents are encoded into 768-dimensional dense vectors using a domain-adapted Chinese medical language model.
    
    \item \textbf{Hierarchical Indexing}: FAISS constructs hierarchical navigable small world (HNSW) graphs \cite{malkov2018efficient} for approximate nearest neighbor search, enabling sub-linear query time complexity.
\end{enumerate}

This indexing strategy ensures both retrieval accuracy and computational efficiency, critical for real-time clinical decision support applications.

\subsubsection{Query Construction and Retrieval}

Given the tongue diagnosis output $T_{\text{tongue}}$ from the visual analysis module and patient metadata $M_{\text{patient}} = (Q, H, P)$, we construct a unified query vector:
\begin{equation}
\mathbf{q} = \text{Embed}(T_{\text{tongue}} \oplus M_{\text{patient}})
\label{eq:query}
\end{equation}
where $\oplus$ denotes information concatenation, and $\text{Embed}(\cdot)$ represents the semantic embedding function mapping text to a dense vector space.

The retrieval module then identifies the $k$ most clinically relevant cases using cosine similarity:
\begin{equation}
R = \text{Retrieve}(\mathbf{q}, k) = \{C_1, C_2, \ldots, C_k\}
\label{eq:retrieve}
\end{equation}
where each retrieved case $C_i$ contains comprehensive diagnostic and treatment information:
\begin{equation}
C_i = \left( T_{\text{tongue}}^{(i)}, Q^{(i)}, H^{(i)}, P^{(i)}, S^{(i)}, Rx^{(i)} \right)
\end{equation}
Here, $S^{(i)}$ represents the syndrome differentiation and $Rx^{(i)}$ denotes the prescribed herbal formula for case $i$.
\subsubsection{Context-Aware Diagnostic Generation}

The final stage of the prescription generator synthesizes the retrieved clinical cases with the current patient's information to generate comprehensive diagnostic outputs. Following the prompt shown in Fig.\ref{prompt} (b), the Qwen3-based reasoning engine processes an evidence-augmented context formed by combining the retrieved cases $R$ with the patient's metadata $M_{\text{patient}}$ and tongue diagnosis $T_{\text{tongue}}$:
\begin{equation}
\text{Output} = \text{Qwen3}(M_{\text{patient}}, T_{\text{tongue}}, R)
\label{eq:generation}
\end{equation}

The system generates four key clinical outputs: (1) syndrome differentiation identifying the underlying TCM patterns, (2) primary diagnosis in biomedical terms, (3) herbal prescription with specific herb combinations and dosages, and (4) clinical reasoning explaining the diagnostic logic and treatment rationale. This comprehensive output format ensures both clinical actionability and interpretability, facilitating physician review and patient understanding.
\begin{algorithm}[t]
\caption{TDEU Score Computation}
\label{alg:tongue-eval}
\begin{algorithmic}[1] 
    \Require Predicted report $\hat{T}$ and reference report $T^*$
    \Ensure Overall similarity score $s$ and per-category similarity dictionary
    \Function{Compute Tongue Eval}{$\hat{T}$, $T^*$}
        \State cfg $\gets$ LoadConfig("token\_config.json")
        \State ts\_pred $\gets$ cfg.extract($\hat{T}$)
        \State ts\_lab $\gets$ cfg.extract($T^*$)
        \State cats\_pred $\gets$ ts\_pred.to\_categories(cfg.category\_map)
        \State cats\_lab $\gets$ ts\_lab.to\_categories(cfg.category\_map)
        \State total $\gets$ 0.0, total\_w $\gets$ 0.0
        \State $\mathcal{C} \gets \{\texttt{TONGUE}, \texttt{COAT}, \texttt{LOCATION}, \texttt{OTHER}\}$ \Comment{Define category set}
        \For{$c \in \mathcal{C}$}
            \State sim $\gets$ \textsc{PairwiseSimilarity}(cats\_pred[$c$], cats\_lab[$c$], cfg)
            \State weight $\gets$ cfg.weights[$c$]
            \State total $\gets$ total + sim $\times$ weight
            \State total\_w $\gets$ total\_w + weight
            \State details[$c$] $\gets$ sim
        \EndFor
        \State overall $\gets$ total / total\_w
        \State \Return (overall, details)
    \EndFunction
\end{algorithmic}
\end{algorithm}

\subsection{TDEU: TCM Diagnosis Evaluation with Semantic Understanding}
Standard text similarity metrics such as BLEU and ROUGE operate on surface-form 
n-gram overlap, failing to capture clinically meaningful equivalences in TCM 
tongue diagnosis. For example, white coating and  (greasy 
coating) are lexically dissimilar but represent distinct pathological states 
(dampness-phlegm vs. interior cold), while pale red and slightly red are near-synonyms. Similarly, omitting critical attributes such as tongue body color carries greater clinical consequence than missing location descriptors. To address these limitations, we introduce TDEU (\textbf{T}CM \textbf{D}iagnosis \textbf{E}valuation with Semantic \textbf{U}nderstanding), a domain-aware metric that incorporates synonym recognition, cross-attribute semantic similarity, and clinical importance weighting. The overall procedure is summarized in Algorithm~\ref{alg:tongue-eval}, which consists of the following components:

\subsubsection{Token Decomposition and Categorization}

Given a predicted tongue report $\hat{T}$ and reference report $T^*$, we 
parse both into atomic attribute tokens using predefined pattern-matching 
rules. Each token is assigned to one of four categories:
\begin{itemize}
\item \texttt{TONGUE}: tongue body color and shape 
\item \texttt{COAT}: coating color, thickness, and texture
\item \texttt{LOCATION}: anatomical regions 
\item \texttt{OTHER}: supplementary descriptors
\end{itemize}

Let $P_c$ and $L_c$ denote the predicted and reference token lists for 
category $c \in \{\texttt{TONGUE}, \texttt{COAT}, \texttt{LOCATION}\}$.

\subsubsection{Semantic Similarity and Optimal Matching}  

For each category $c$, we compute pairwise semantic similarities between 
tokens in $P_c$ and $L_c$. Each token follows the format \texttt{PREFIX\_VALUE} 
(e.g., \texttt{COLOR\_red}, \texttt{THICK\_thin}). Similarity $S(p, l)$ is 
defined through a hierarchical matching scheme:

\begin{equation}
S(p, l) = 
\begin{cases}
1.0 & \text{if exact match} \\
\text{Syn}_{\pi(p)}(\nu(p), \nu(l)) & \text{if } \pi(p) = \pi(l) \\
\text{Cross}(p, l) & \text{if } \pi(p) \neq \pi(l) \\
0.0 & \text{otherwise}
\end{cases}
\end{equation}

where $\text{Syn}_c(v_1, v_2)$ denotes the expert-curated synonym similarity 
score for values $v_1, v_2$ within category $c$. These scores range from 0.6 
to 0.9 based on clinical equivalence (e.g., $\text{Syn}_{\text{COLOR}}(\text{pale}, \text{pale\_white}) = 0.9$, 
$\text{Syn}_{\text{COLOR}}(\text{red}, \text{dark\_red}) = 0.6$). The cross-category 
similarity $\text{Cross}(p, l)$ captures semantic correlations between 
attributes from different categories (e.g., $\text{Cross}(\text{SHAPE\_swollen}, \text{THICK\_thick}) = 0.3$), 
with scores ranging from 0.2 to 0.9 based on TCM diagnostic principles.

We apply the Hungarian algorithm to find the optimal one-to-one matching 
$M_c \subseteq P_c \times L_c$ that maximizes total similarity:
\begin{equation}
M_c = \arg\max_{M} \sum_{(p,l) \in M} S(p, l).
\end{equation}

The category-level similarity score incorporates a partial matching bonus 
to reward correctly identified attributes even when the prediction is incomplete:
\begin{equation}
\text{TokenListSim}_c = \max\left(\text{BaseScore}_c, \text{BonusScore}_c\right)
\end{equation}
where
\begin{equation}
\text{BaseScore}_c = \frac{\sum_{(p,l) \in M_c} S(p,l)}{\max(|P_c|, |L_c|)}
\end{equation}
and
\begin{equation}
\text{BonusScore}_c = 0.3 \times \frac{|\{(p,l) \in M_c : S(p,l) > 0\}|}{\max(|P_c|, |L_c|)}.
\end{equation}

This ensures that predictions with partial correctness receive a minimum score 
proportional to the number of matched tokens, preventing over-penalization of 
incomplete but clinically relevant predictions.

\subsubsection{Clinical Importance Weighting}
We evaluate four categories: \texttt{TONGUE}, \texttt{COAT}, \texttt{LOCATION}, and \texttt{OTHER} (auxiliary descriptors). Through consultation with senior TCM practitioners, we let $w_{\text{TONGUE}}{=}2.0$, $w_{\text{COAT}}{=}1.5$, $w_{\text{LOCATION}}{=}1.0$, and $w_{\text{OTHER}}{=}0.5$.
Define the category set $\mathcal{C}=\{\mathrm{TONGUE},\mathrm{COAT},\mathrm{LOCATION},\mathrm{OTHER}\}$.
The overall score is a weighted average:
\begin{equation}
\mathrm{TDEU} \;=\;
\frac{\sum_{c \in \mathcal{C}} w_c \cdot \mathrm{TokenListSim}_c}{\sum_{c \in \mathcal{C}} w_c}.
\end{equation}

In Table~\ref{tab:combined_all}, we report the sub-scores for \texttt{TONGUE}, \texttt{COAT}, and \texttt{LOCATION} to improve readability. The \textbf{Overall} score follows the four-category weighted aggregation scheme described above, which also includes \texttt{OTHER}. In our implementation, the \texttt{OTHER} category represents auxiliary descriptors such as teeth-mark impressions, surface cracks, and moisture-related adjectives that are not covered by the other three categories.

\section{Results}

\subsection{Dataset Construction: MedTCM}

To address the critical scarcity of large-scale multimodal TCM datasets, we constructed MedTCM through collaboration with four leading TCM hospitals in China. The study protocol received ethical approval from the Institutional Review Boards of all participating institutions (ethical approval number: 2025(S) No. 029). All patient data were rigorously de-identified to protect privacy, with all personally identifiable information (PII) systematically removed.

\subsubsection{Multi-Center Data Collection Strategy}

\begin{figure}[!t]
\centering
\includegraphics[width=\columnwidth]{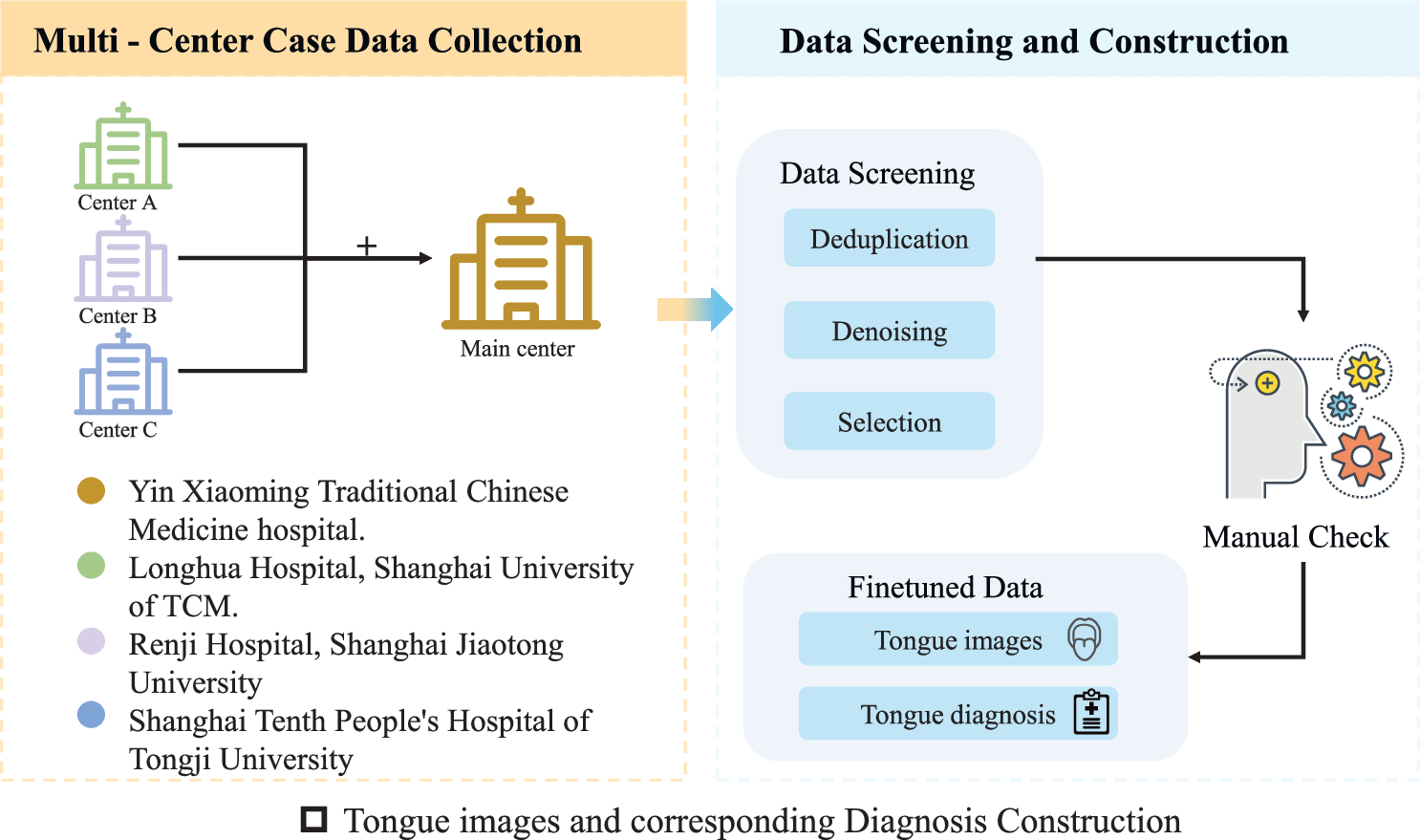}
\caption{MedTCM multi-center data construction and curation pipeline. High-quality tongue images are paired with de-identified clinical records and expert annotations.}
\label{fig:medtcm}
\end{figure}

The data collection framework, illustrated in Fig.~\ref{fig:medtcm}, was structured around a primary institution leading data standardization, complemented by three partner hospitals contributing diverse clinical cases. This multi-center strategy was crucial for ensuring data diversity and mitigating biases inherent in single-center collections \cite{liu2025diagnosing,lin2024machine}.

The collaborative effort yielded an extensive corpus of 124,593 anonymized patient records spanning the period from April 2021 to June 2025. Each record contains comprehensive clinical information including:
\begin{itemize}
    \item Chief complaint and present illness history
    \item Tongue diagnosis (visual and textual description)
    \item Pulse diagnosis findings
    \item TCM syndrome differentiation
    \item Herbal prescription with specific formulations
\end{itemize}

From this corpus, we curated a high-resolution multimodal subset of 2,805 tongue images, each meticulously paired with its corresponding expert-annotated diagnostic description and complete medical record. To the best of our knowledge, MedTCM represents the first large-scale, open-source dataset integrating visual tongue imagery with comprehensive clinical records and expert annotations.

\begin{table}[t]
  \centering
  \caption{Detailed Characteristics and Statistical Distribution of the MedTCM Dataset.}
  \label{tab:dataset_statistics}
  \begingroup
  \footnotesize                           
  \setlength{\tabcolsep}{3pt}             
  \renewcommand{\arraystretch}{1.05}      
  \setlength{\aboverulesep}{0pt}          
  \setlength{\belowrulesep}{0pt}          
  \captionsetup{skip=4pt}                

  \begin{tabularx}{\linewidth}{@{}>{\bfseries}p{0.20\linewidth} p{0.22\linewidth} p{0.25\linewidth} Y@{}}
    \toprule
    Category & Characteristic & Metric / Sub-category & Value / Distribution \\
    \midrule
    \multirow{3}{*}{\makecell[l]{Overall\\Statistics}}
      & Cases Count & Records & 124{,}593 cases \\
      & Tongue Subset     & Image--Diagnosis  & 2{,}805 pairs \\
      & Time period     & -- & 04.2021 -- 06.2025\\
    \midrule
    \multirow{4}{*}{\makecell[l]{Patient\\Demographics}}

      & \multirow{2}{*}{Age (years)} & Mean $\pm$ SD       & 37.6 $\pm$ 15.2 \\
      &                               & Range    & 1--96 \\
      \cmidrule(l){2-4}
      & \multirow{2}{*}{Gender}       & Male                & 37369 (30\%) \\
      &                               & Female              & 87,224 (70\%) \\
\midrule
  \multirow{2}{*}{\makecell[l]{Prescriptions}}
   & \multirow{2}{*}{Categories} & syndrome diff.   & 855 types\\
  &                                     & Tongue Diagnosis & 7294 types \\
  
    \bottomrule
  \end{tabularx}
  \endgroup
\end{table}
\subsubsection{Dataset Statistics and Characteristics}
As shown in Table~\ref{tab:dataset_statistics}, the dataset exhibits rich demographic diversity and comprehensive syndrome coverage. The tongue image subset (2,805 pairs) was selected to ensure high image quality. Expert TCM physicians verified all image-diagnosis pairs for consistency and clinical accuracy. The dataset will be continuously updated and released as open source to facilitate future research in AI-assisted TCM diagnosis.

\medskip



\begin{table*}[t]
  \centering
  \caption{Combined results: baselines, ViTCM-LLM ablations, and MMIR-TCM (ours). Missing values are shown as “–”.}
  \label{tab:combined_all}
  \setlength{\tabcolsep}{6pt}
  \renewcommand{\arraystretch}{1.12}
  \small
  \resizebox{\textwidth}{!}{%
  \begin{tabular}{l c c c | cccc | cccc}
    \toprule
    \multirow{2}{*}{\textbf{Model / Setting}} & \multirow{2}{*}{\textbf{Params}} & \multirow{2}{*}{\textbf{LoRA $r$}} & \multirow{2}{*}{\textbf{Epochs}} & \multicolumn{4}{c|}{\textbf{BLEU / ROUGE}} & \multicolumn{4}{c}{\textbf{TDEU}} \\
    \cmidrule(lr){5-8} \cmidrule(lr){9-12}
    & & & & \textbf{BLEU-4} & \textbf{R-1} & \textbf{R-2} & \textbf{R-L} & \textbf{Overall} & \textbf{Tongue} & \textbf{Coat} & \textbf{Location} \\
    \midrule
    \multicolumn{12}{l}{\textit{Baselines (zero-shot)}} \\
    Grok\textendash 2\textendash Vision\textendash 1212 & --   & -- & -- & 26.40 & 22.12 & 22.08 & 26.43 & 0.338 & 0.262 & 0.404 & 0.367 \\
    LLaMA4\textendash scout 109B                       & 109B & -- & -- & 25.68 & 22.45 & 22.43 & 25.70 & 0.336 & 0.254 & 0.412 & 0.356 \\
    Gemini\textendash 2.5\textendash Flash             & --   & -- & -- & 24.43 & 21.37 & 21.33 & 24.46 & 0.353 & 0.271 & 0.421 & 0.372 \\
    \midrule
    \multicolumn{12}{l}{\textit{ViTCM\textendash LLM (Qwen2.5\textendash VL 32B)}} \\
    E1 Zero\textendash shot               & 32B & -- & -- & 24.08 & --    & --    & --    & 0.3538 & --    & --    & --    \\
    E2 Language\textendash only           & 32B & 16 & 3  & 26.74 & --    & --    & --    & 0.2698 & --    & --    & --    \\
    E3 Vision+Projector                   & 32B & 16 & 3  & 35.82 & --    & --    & --    & 0.3610 & --    & --    & --    \\
    E4 Full ($r{=}16$, 3ep)              & 32B & 16 & 3  & 37.94 & --    & --    & --    & 0.3648 & --    & --    & --    \\
    E4 Full ($r{=}64$, 3ep)              & 32B & 64 & 3  & 39.57 & --    & --    & --    & 0.3915 & --    & --    & --    \\
    E4 Full ($r{=}64$, 10ep)             & 32B & 64 & 10 & 43.57 & --    & --    & --    & 0.5858 & --    & --    & --    \\
    E4 Full ($r{=}64$, 20ep)             & 32B & 64 & 20 & 44.07 & --    & --    & --    & 0.6150 & --    & --    & --    \\
    \midrule
    \multicolumn{12}{l}{\textbf{MMIR\textendash TCM (ours; Qwen3\textendash VL 30B + LoRA)}} \\
    Original Qwen3\textendash VL 30B (untuned) & 30B & -- & -- & 33.98 & 28.42 & 28.23 & 34.05 & 0.289 & 0.187 & 0.288 & 0.141 \\
    Raw train + Raw   & 30B & 64 & 10 & 83.20 & 76.24 & 76.06 & 83.62 & 0.601 & 0.442 & 0.673 & 0.611 \\
    Raw train + Mask  & 30B & 64 & 10 & 83.11 & 75.96 & 75.57 & 83.34 & 0.612 & 0.472 & 0.683 & 0.590 \\
    \textbf{Mask train + Raw}  & 30B & 64 & 10 & 83.22 & 76.06 & 75.69 & 83.45 & 0.617 & \textbf{0.478} & 0.693 & 0.592 \\
    \textbf{Mask train + Mask} & 30B & 64 & 10 & \textbf{83.57} & \textbf{77.30} & \textbf{76.77} & \textbf{84.30} & \textbf{0.627} & 0.473 & 0.693 & \textbf{0.649} \\
    Raw train + Raw   & 30B & 64 & 3  & 81.72 & 74.42 & 73.88 & 82.20 & 0.580 & 0.385 & 0.682 & 0.607 \\
    \textbf{Mask train + Mask} & 30B & 64 & 3  & 83.09 & 76.50 & 76.28 & 83.73 & 0.611 & 0.439 & \textbf{0.698} & 0.632 \\
    \bottomrule
  \end{tabular}
  }
\end{table*}

\subsection{Tongue Extractor Performance}
\label{subsec:seg_performance}

\textbf{Rationale and prior validation.}
We adopt Memory-SAM as a \emph{training-free inference} pipeline that avoids task-specific model fine-tuning.
It requires only a \textbf{small gallery of annotated exemplars} (binary masks) to convert exemplar correspondences into explicit foreground/background point prompts for SAM2, rather than large-scale pixel-level labeling of the entire dataset.
This “few-annotation” setup drastically reduces annotation cost while retaining robustness under acquisition shifts.\cite{chae2025memory}

\begin{figure}[!t]
  \centering
  \includegraphics[width=\linewidth]{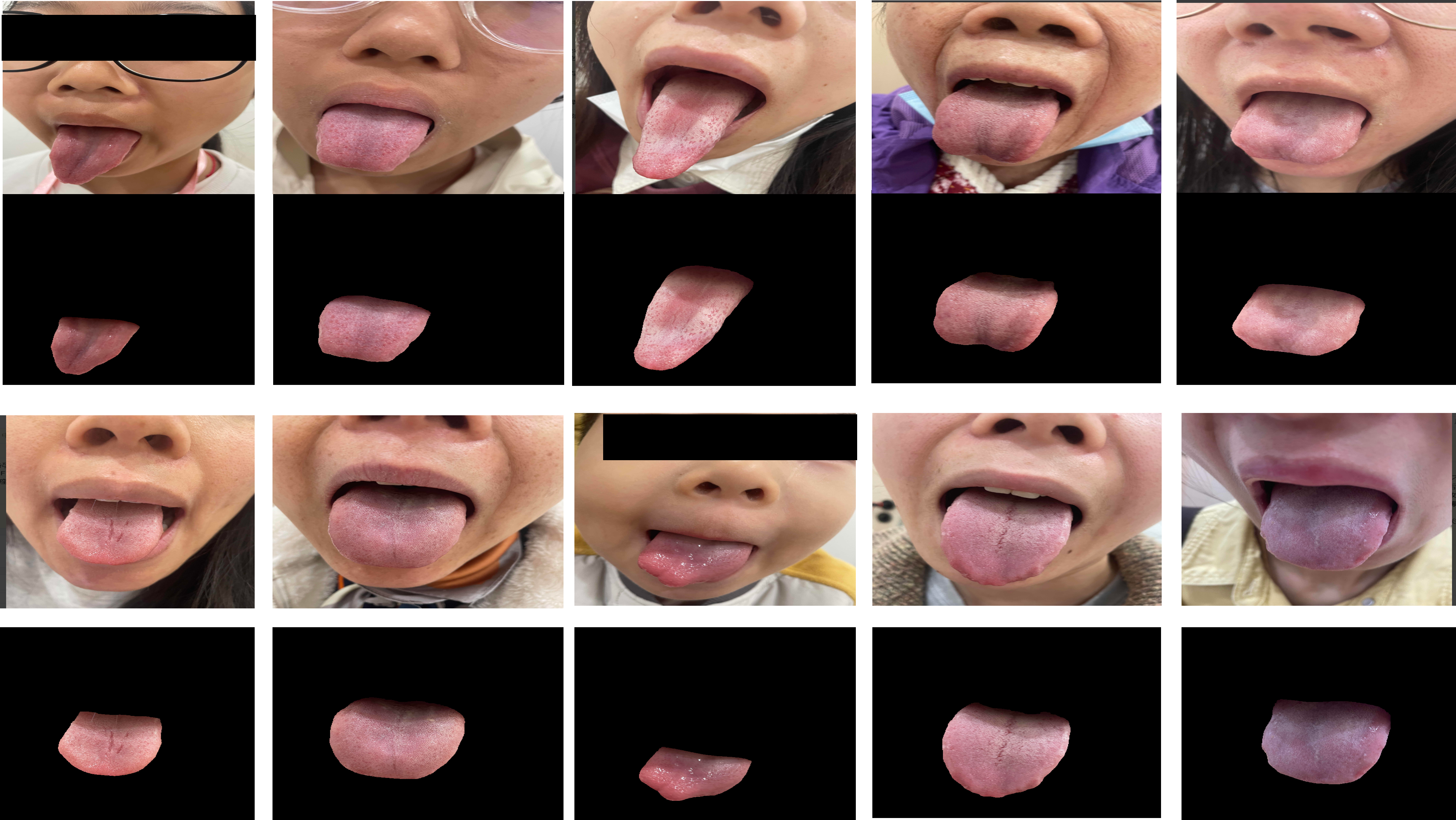}
  \caption{Background removal and ROI standardization with Memory-SAM. For diverse in-the-wild inputs, extracted tongue masks produce clean foreground crops that reduce illumination and background variance, stabilizing the interface for report and prescription generation.}
  \label{fig:bg_removal}
\end{figure}

\medskip
\textbf{Operational yield on corpus.}
Across the full dataset of 2,805 images, Memory-SAM successfully generated clinically usable masks for \textbf{2,780} cases as shown in Fig.\ref{fig:bg_removal} (\textbf{99.1\%} success rate). The remaining 25 cases were manually corrected before they were entered into the training data, mainly due to low contrast between the tongue and background or under-exposed images.

\subsection{Tongue Diagnosis Generator Performance}
\label{subsec:report_perf}

\textbf{Setup.}
We evaluate a tongue report generator built on \textbf{Qwen3\textendash VL\textendash 30B} under six settings that vary the input type for training and evaluation (raw vs.\ segmented) and the number of fine\textendash tuning epochs (3, 10).
All in\textendash house runs use the same LoRA configuration (\textbf{rank $r{=}64$, $\alpha{=}128$, dropout $=0$}) and learning rate \textbf{$5{\times}10^{-5}$}.

The \emph{Original Qwen3\textendash VL\ 30B} model without domain fine\textendash tuning serves as an internal baseline.
For external reference, we additionally report three public MLLMs (\emph{Grok\textendash 2\textendash Vision\textendash 1212}, \emph{LLaMA4\textendash scout 109B}, \emph{Gemini\textendash 2.5\textendash Flash}) evaluated \emph{zero\textendash shot}.
Metrics include \textbf{BLEU\textendash 4}, \textbf{ROUGE}, and domain\textendash aware clinical fidelity \textbf{TDEU} (Overall, Tongue, Coat, Location), summarized in Table~\ref{tab:combined_all}.

\textbf{Results.}
Using segmented inputs consistently at both training and evaluation (\textbf{Mask$\to$Mask}, 10 epochs) yields the best text metrics (\textbf{BLEU\textendash 4 83.57}, \textbf{ROUGE\textendash 1/2/L = 77.30/76.77/84.30}).
Training and evaluating on raw images (\emph{Raw$\to$Raw}, 10 epochs) is competitive (BLEU\textendash 4 83.20; ROUGE\textendash L 83.62).
On the domain\textendash aware TDEU, \textbf{Mask$\to$Mask} (10 epochs) attains the highest \textbf{Overall 0.627}, with sub\textendash category scores indicating that \emph{Coat} and \emph{Location} are comparatively easier to match than \emph{Tongue}.
Zero\textendash shot public MLLMs remain far below the domain\textendash adapted models (BLEU\textendash 4 mid\textendash 20s; TDEU Overall $\approx$ 0.34–0.35).
For a detailed analysis of how input standardization with a tongue extractor affects performance across training/evaluation stages, see the ablation in Sec.~\ref{subsec:abl_tongue_extractor}.

\subsection{Prescription Generator Performance}
\textbf{Setup.}
For our quantitative evaluation, we constructed a held-out test set by randomly sampling 1,151 cases from MedTCM. To ensure comprehensive coverage of diagnostic scenarios, this sample was stratified to include representative cases from all major syndrome differentiation types. For each case, we extracted the patient's visual tongue diagnosis results along with other clinical metadata ${M}_{\text{patient}}$ to serve as the input for the models. All experiments were conducted with a fixed random seed of 42. The publicly released model \textit{Qwen/Qwen3-8B}, serving as the LLM backbone, was deployed on 8 \textbf{NVIDIA RTX 3090 GPUs} (24 GB). Our implementation integrates a sophisticated retrieval and indexing method.The retrieval component utilizes the LangChain framework with FAISS as the underlying vector database for efficient similarity search. Case documents are indexed using HNSW graphs within FAISS, optimized for fast approximate nearest neighbor search. Generation uses nucleus sampling with p=0.9, temperature=0.7, and top-k=50 to balance creativity and coherence. 
\begin{figure*}[!t]
\centering
\includegraphics[width=0.9\textwidth]{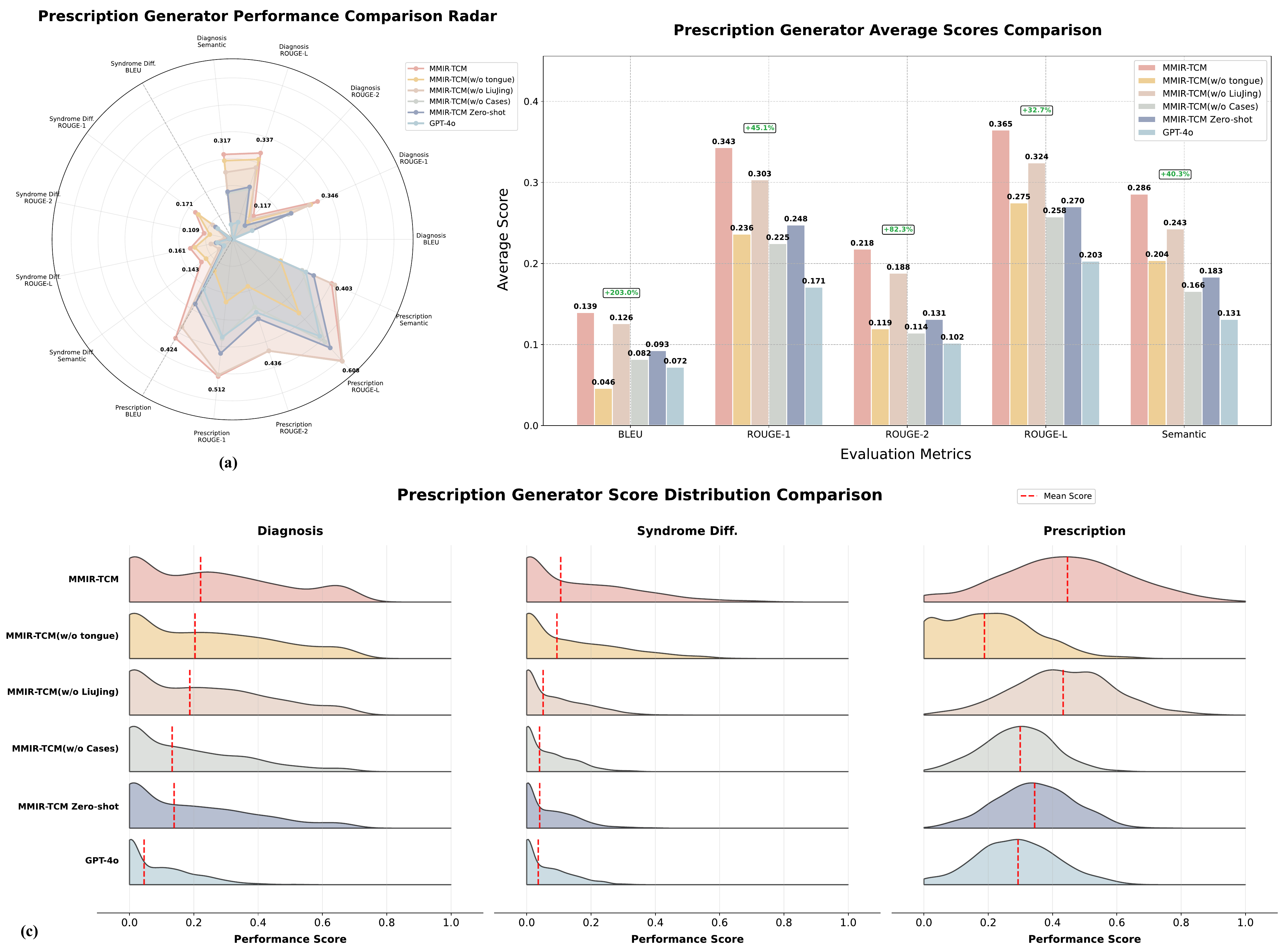}
\caption{Overall performance comparison. (a)~Radar chart shows MMIR-TCM’s consistent superiority. (b)~Average results across tasks. (c)~Score distributions highlight higher median and stability.}
\label{fig:Prescription_Overall}
\end{figure*}

\textbf{Results.}
A qualitative examination of Fig.~\ref{fig:Prescription_Overall}(a) reveals that the performance polygon of MMIR-TCM consistently encloses those of all baseline models, underscoring its robust superiority across the entire evaluation spectrum. To further quantify this superiority, Fig.~\ref{fig:Prescription_Overall}(b) presents the averaged results across five evaluation metrics and three representative clinical tasks. MMIR-TCM achieves the highest overall performance, reaching the 5 scores of 0.142, 0.343, 0.220, 0.369, and 0.288, respectively—corresponding to relative gains of +209.6\%, +45.2\%, +84.6\%, +34.1\%, and +41.3\% over the ablated variant without tongue diagnosis (MMIR-TCM w/o tongue). These improvements highlight the critical contribution of tongue diagnosis in enhancing the semantic fidelity and diagnostic precision of prescription generation. Fig.~\ref{fig:Prescription_Overall}(c) further illustrates the distribution of aggregated performance scores, where MMIR-TCM exhibits a higher median and a more compact, right-skewed density, indicating stronger stability and generalization in clinical prescription reasoning.

\subsection{Ablation Study}
\subsubsection{Ablation on Tongue Extractor}
\label{subsec:abl_tongue_extractor}
\begin{figure*}[!ht]
\centering
\includegraphics[width=0.9\textwidth]{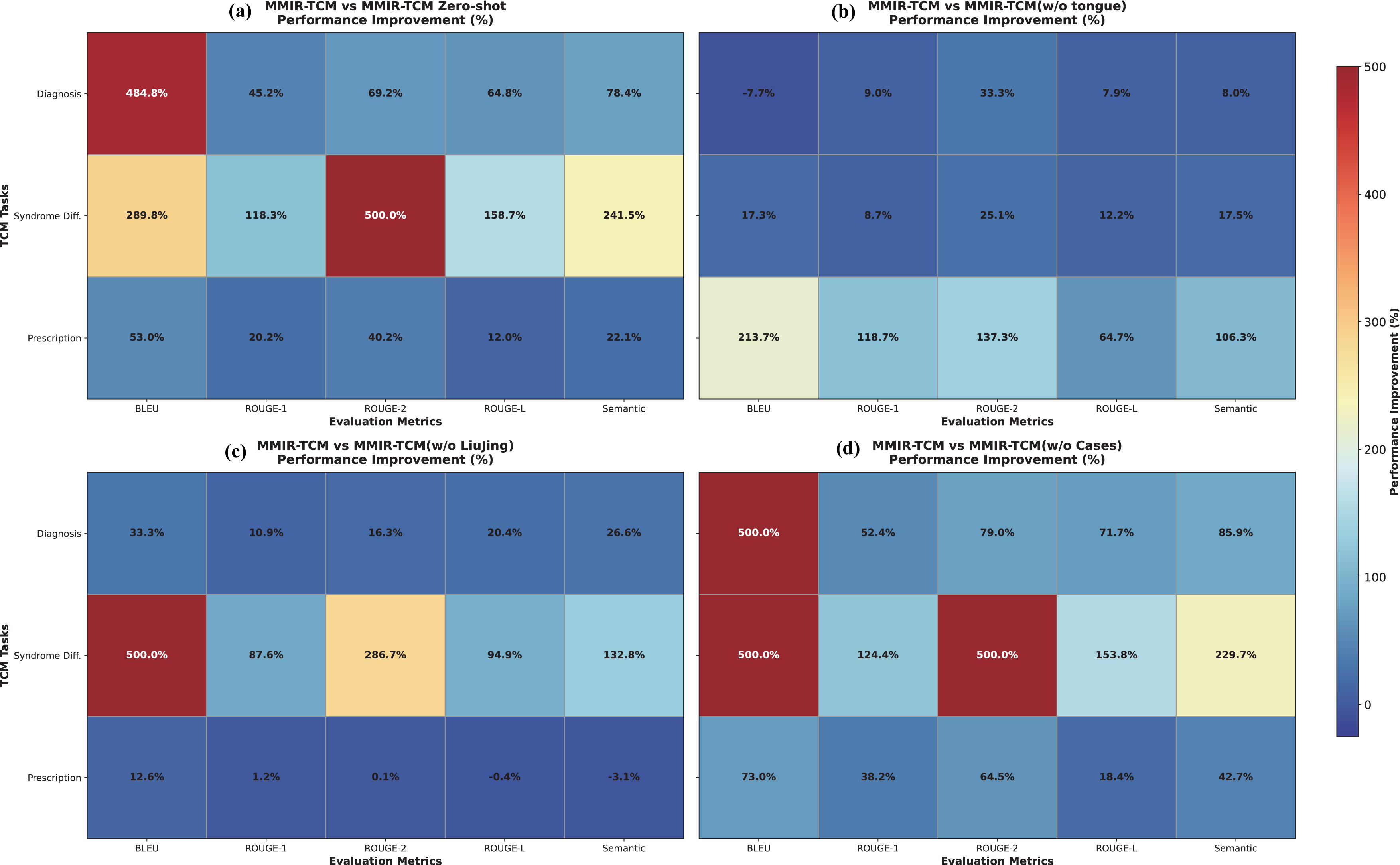}
\caption{Heatmap  Visualization  of  Performance  Gains from MMIR-TCM's Prescription Generator Framework.}
\label{fig:RAG_Heatmaps}
\end{figure*}
In deep learning-based tongue diagnosis and medical imaging, it is well-established that mask segmentation (background removal) or ROI standardization enhances model performance and stability. This is generally attributed to two factors: (i) suppressing non-medical sources of variance such as background clutter, illumination, and skin tone, and (ii) guiding downstream modules to focus on the semantically meaningful region (in this case, the tongue). However, while these benefits are well-documented for traditional computer vision tasks like classification, segmentation, and detection, or in multimodal contexts where text is secondary, their effects have not been systematically validated in scenarios centered on language generation, such as generating tongue diagnosis reports with MLLMs.

This ablation study is designed to address precisely this gap. Specifically, we evaluate the extent to which input standardization via mask segmentation improves text quality and clinical fidelity in MLLM-based tongue report generation. Furthermore, we assess whether the magnitude of this improvement is sufficient to justify its adoption from a practical and operational standpoint.

Evaluation relied on BLEU-4 and ROUGE for text quality, and on the clinically aware fidelity metric TDEU (Overall, Tongue, Coat, Location) (Table~\ref{tab:combined_all}). Using segmented inputs consistently for both training and evaluation (\(\text{Mask}\,\textrightarrow\,\text{Mask}\)) achieved the best performance at 10\,epochs (BLEU-4\,83.57; ROUGE-1/2/L\,77.30/76.77/84.30; TDEU Overall\,0.627). The fully raw pipeline (\(\text{Raw}\,\textrightarrow\,\text{Raw}\)) was competitive (BLEU-4\,83.20; ROUGE-L\,83.62; TDEU Overall\,0.601), but the \(\text{Mask}\,\textrightarrow\,\text{Mask}\) approach remained consistently superior in both lexical overlap and clinical semantics.

Crossed conditions, such as \(\text{Raw}\,\textrightarrow\,\text{Mask}\) and \(\text{Mask}\,\textrightarrow\,\text{Raw}\), where segmentation was applied at only one stage, produced nearly identical BLEU-4 scores at 10\,epochs ($\approx$\,83.22). However, they yielded modestly higher TDEU Overall scores (0.612 and 0.617, respectively) than the \(\text{Raw}\,\textrightarrow\,\text{Raw}\) pipeline, indicating that standardizing inputs at either the training or evaluation stage improves semantic stability. Even with only 3\,epochs of training, the \(\text{Mask}\,\textrightarrow\,\text{Mask}\) setup (BLEU-4\,83.09; TDEU\,0.611) outperformed the 3-epoch \(\text{Raw}\,\textrightarrow\,\text{Raw}\) model (BLEU-4\,81.72; TDEU\,0.580), demonstrating that background suppression and shape normalization yield immediate performance gains.

In conclusion, mask segmentation provides a modest benefit by reducing variance from background and illumination, slightly improving the stability of report generation. While its impact in a MLLM-based tongue diagnosis model is limited and it cannot be considered an essential component, it is retained in the system to ensure consistent input formatting for downstream modules. In other words, it serves as a helpful preprocessing step that supports the pipeline’s robustness. Achieving higher diagnostic accuracy likely requires strategies beyond input standardization, such as expanding dataset size or exploring more advanced fine-tuning techniques beyond LoRA (e.g., full/partial fine-tuning, vision tower alignment, color calibration, and robust data augmentation).

\subsubsection{Ablation on Fine-Tuning Strategy and Hyperparameters}

\begin{table}[t]
  \centering
  \caption{LoRA fine-tuning scenarios. "LoRA" indicates that new low-rank adapters are trained while the original weights remain frozen.}
  \label{tab:scenarios}
  \setlength{\tabcolsep}{6pt}
  \begin{tabular}{lccc}
    \toprule
    Scenario & Vision Tower & Projector & Language LM \\
    \midrule
    Baseline (E1)            & Frozen & Frozen & Frozen \\
    Language-only (E2)       & Frozen & Frozen & LoRA   \\
    Vision{+}Projector (E3)  & LoRA   & LoRA   & Frozen \\
    Full Model (E4)          & LoRA   & LoRA   & LoRA   \\
    \bottomrule
  \end{tabular}
\end{table}
The model architecture is composed of three main components: (i) a \textbf{vision tower}, (ii) a \textbf{multimodal projector}, and (iii) a \textbf{LLaMA-style language transformer}. LoRA adapters are inserted into the linear layers of both the \textbf{self-attention} modules ($q,k,v,o$) and the \textbf{feed-forward} modules (gate, up, down). The training configuration employs a batch size of~4 with two-step gradient accumulation for \textbf{3 epochs}. \textbf{Early stopping} is applied when the validation loss does not improve for three consecutive epochs.

As summarized in Table~\ref{tab:scenarios}, to separate component-wise contributions and locate an efficiency–performance trade-off, we evaluate three controlled configurations:
\begin{itemize}
  \item \textbf{Language-only (L-only):} LoRA applied exclusively to the language transformer.
  \item \textbf{Vision + Projector (V+Proj):} LoRA applied to the vision tower and multimodal projector.
  \item \textbf{Full:} LoRA applied jointly to all three components.
\end{itemize}

To further investigate the balance between representational capacity and overfitting, we vary both the \textbf{LoRA rank} ($r$) and the \textbf{training epochs}. The results, presented in Table~\ref{tab:combined_all}, show the following key observations:

\begin{itemize}
  \item \textbf{Module contributions.} \textit{V+Proj} yields larger gains than \textit{L-only} on both BLEU/TDEU, and \textit{Full} achieves the best overall performance, indicating that visual domain grounding is more influential than language-only adaptation.
  \item \textbf{Capacity \& schedule.} Increasing $r$ from $16$ to $64$ and extending training from 3 to 10 epochs jointly improve BLEU-4 and TDEU. However, moving from 10 to 20 epochs leads to only modest BLEU-4 gains while TDEU shows diminishing returns, suggesting early signs of overfitting (language-dominated drift).
\end{itemize}

\begin{figure*}[!t]
\includegraphics[width=0.9\textwidth]{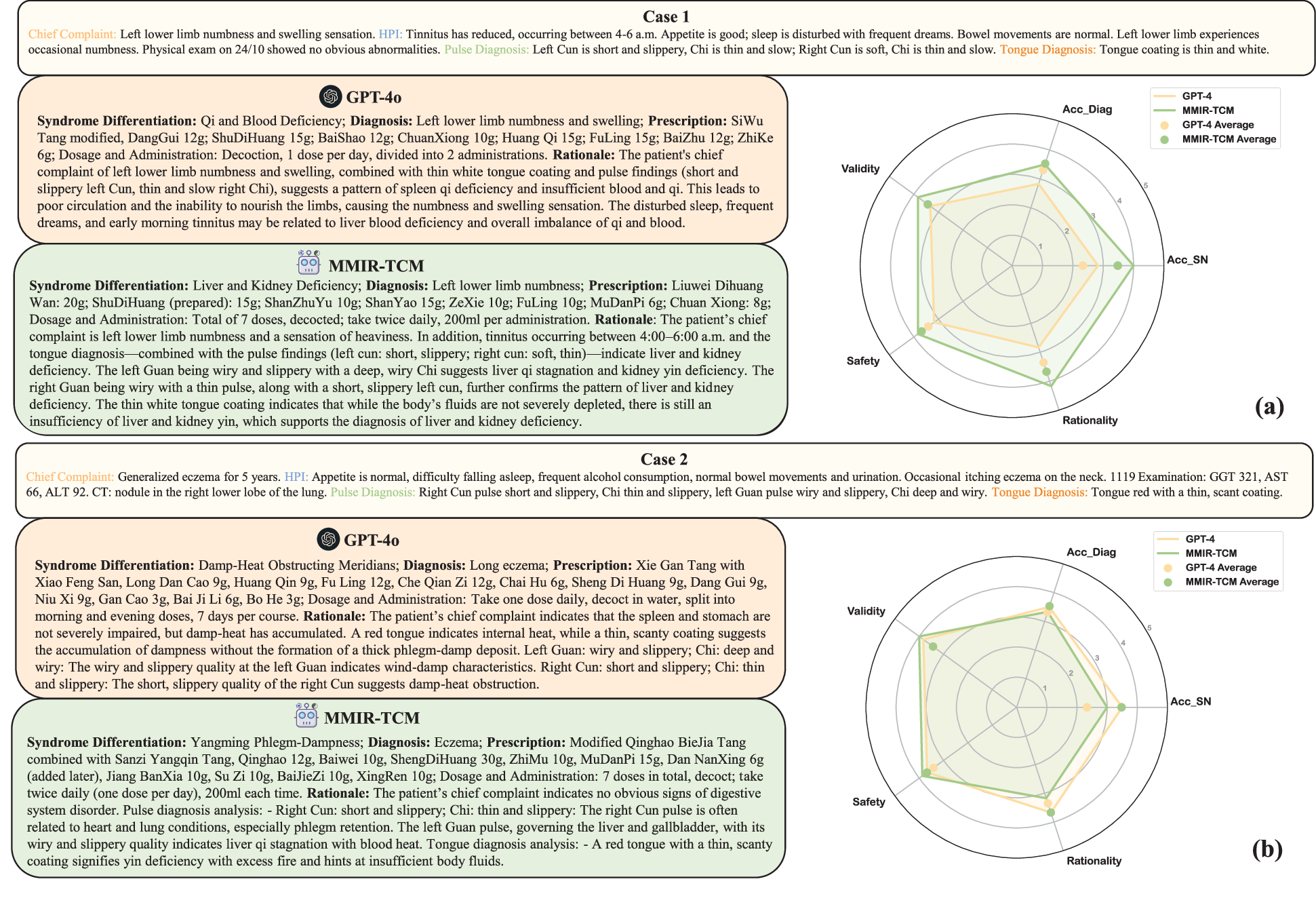}
\caption{Comparison of MMIR-TCM and GPT-4o prescriptions. The left panels show two representative cases: (a) a case where MMIR-TCM outperforms GPT-4o, and (b) a case favoring GPT-4o. The radar plot on the right summarizes physicians’ average evaluation scores across all cases, highlighting overall model performance.} \label{User_preference_result}
\end{figure*}
\subsubsection{Ablation on Prescription Generator}
To validate the effectiveness and contribution of each key component of our proposed MMIR-TCM’s RAG framework, we conducted a series of ablation studies. We systematically removed specific modules from the full model and evaluated the performance degradation. The results, visualized as heatmaps in Fig.~\ref{fig:RAG_Heatmaps}, compare the full MMIR-TCM model against four ablated variants across three core TCM tasks (Diagnosis, Syndrome Differentiation, and Prescription) using five evaluation metrics. For clarity in visualization, performance improvements exceeding 500\% are capped at 500\% in the heatmaps.

The first comparison reveals a substantial performance gap against the zero-shot baseline (Fig.~\ref{fig:RAG_Heatmaps}a), with improvements reaching 484.8\% (BLEU for Diagnosis) and 500.0\% (ROUGE-2 for Syndrome Differentiation). Component-specific ablations further highlight their individual contributions. Removing the clinical cases memory (\texttt{w/o Cases}, Fig.~\ref{fig:RAG_Heatmaps}d) caused the most significant degradation, with performance dropping by 500.0\% on both Diagnosis and Syndrome Differentiation. Similarly, ablating the LiuJing theory memory (\texttt{w/o LiuJing}, Fig.~\ref{fig:RAG_Heatmaps}c) severely compromised Syndrome Differentiation (500.0\% drop in BLEU score). The removal of tongue diagnosis (\texttt{w/o 
tongue}, Fig.~\ref{fig:RAG_Heatmaps}b) also led to a notable decline, particularly impacting the Prescription task with a 213.7\% drop in BLEU.

These findings demonstrate that the clinical cases memory, LiuJing theory memory, and tongue diagnosis are all integral yet complementary components, each providing distinct and non-redundant contributions to the model’s overall efficacy in complex TCM reasoning tasks. This validates the architectural design choices of the proposed MMIR-TCM framework.

\subsection{User Preference Study.}
\subsubsection{Overall Ratings}
We conducted a blinded preference study with 12 experienced TCM clinicians comparing MMIR-TCM and GPT-4o on 20 random cases (240 evaluations). Outputs were scored from 1 to 5 on five dimensions: Acc\_SN, Acc\_Diag, Validity, Safety, and Rationality.
\begin{figure*}[!t]
\includegraphics[width=0.9\textwidth]{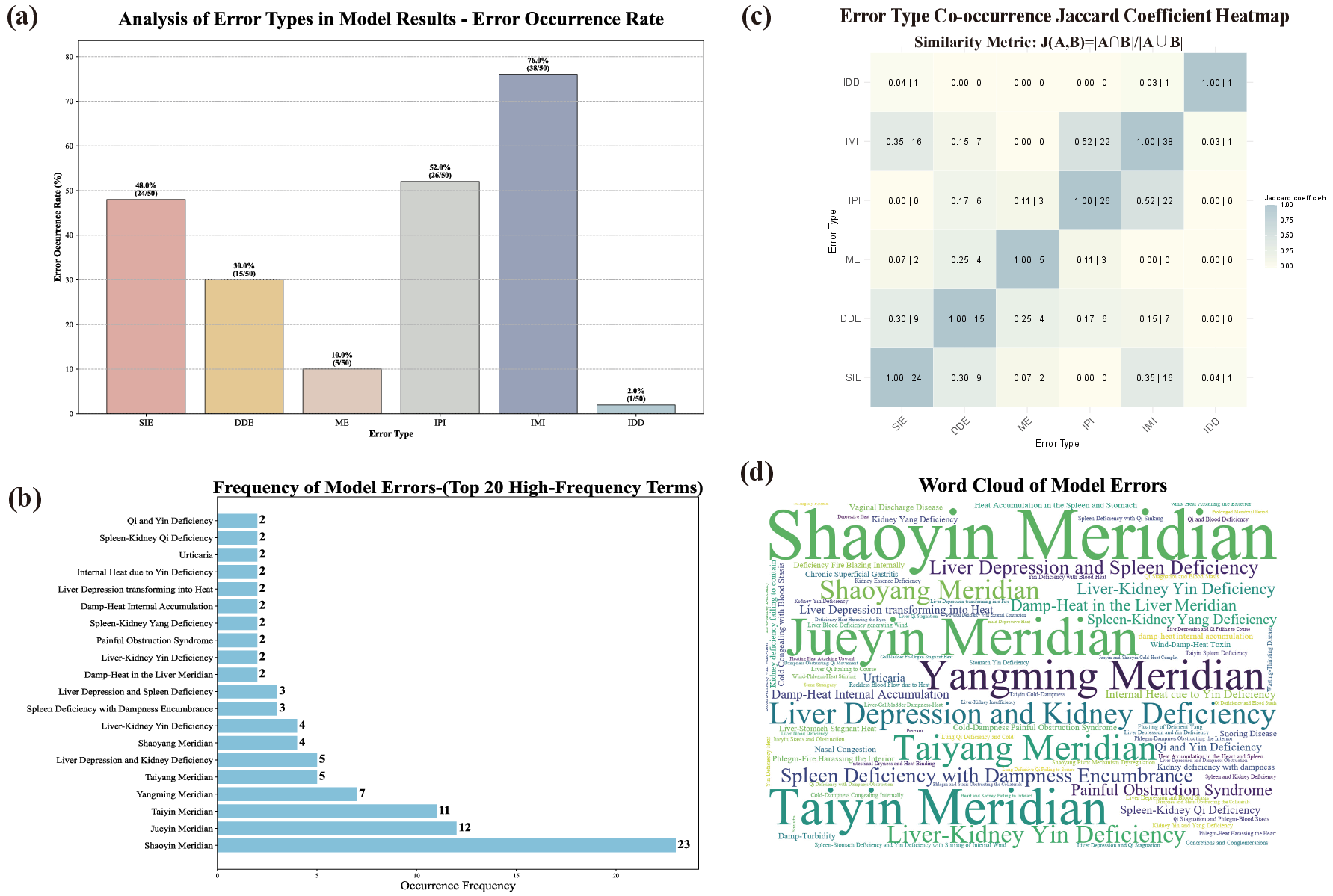}
\caption{Statistical analysis of representative  erroneous cases in MMIR-TCM}
\label{fig:Failures_Analysis}
\end{figure*}
Aggregated ratings (Fig. \ref{User_preference_result}) show a consistent clinician preference for MMIR-TCM, with feedback indicating better alignment with TCM reasoning. Case 1 strongly favored MMIR-TCM. In Case 2, GPT-4o scored slightly higher on the five metrics, but clinicians still preferred MMIR-TCM due to stronger perceived safety and reliability, emphasizing that diagnostic correctness and prescription safety are primary priorities.

\subsubsection{Challenges and Limitations}
To quantify real-world accuracy, we performed manual error analysis against formal TCM hospital records. Five experts reviewed 2,362 generated cases and re-evaluated a random 50\% subset (1,181) for disease diagnosis, meridian attribution, and syndrome consistency. Inconsistencies were marked as incomplete or incorrect, and 50 representative error cases were analyzed in six categories: SIE, DDE, ME, IPI, IMI, and IDD.

Frequency analysis (Fig. \ref{fig:Failures_Analysis}(a)) showed IMI as the dominant error (76.0\%, 38/50), followed by IPI (52.0\%, 26/50), indicating a priority to improve meridian and syndrome completeness. High-frequency terms (Fig. \ref{fig:Failures_Analysis}(b)) were concentrated in Shaoyin (23) and Jueyin (12), and in complex syndrome combinations (e.g., Liver Depression and Kidney Deficiency; Liver--Kidney Yin Deficiency), suggesting limited integration of multifactorial TCM cues.

Co-occurrence analysis (Fig. \ref{fig:Failures_Analysis}(c)) showed strong IPI--IMI coupling ($>$20 times, Jaccard = 0.52, co-occurrence $>$50\%), moderate SIE--IMI coupling (16), and near-independence of ME--SIE (Jaccard = 0.07). This suggests "Error" and "Incomplete" types may arise from different reasoning biases and should be optimized separately. The word cloud (Fig. \ref{fig:Failures_Analysis}(d)) confirms concentration in meridian--syndrome differentiation and complex combinations; clinically this indicates weakness in disease--meridian linking and deficiency/viscera reasoning, while technically it likely reflects terminology bias and limited contextual TCM semantic understanding.

\subsection{Clinical Relevance Study}
To assess translational potential, we examined practical applicability in real-world clinical settings. The framework’s core value is improved diagnostic objectivity, efficiency, and safety. Standardized tongue-pattern analysis provides quantitative visual references, while the RAG mechanism enables rapid retrieval of similar cases and preliminary diagnostic reports. These outputs are grounded in an authentic clinical knowledge base, reducing “hallucinations” and improving reliability.

Nonetheless, error analysis in Fig.~\ref{fig:Failures_Analysis} highlights key challenges. Among erroneous cases, 76\% of IMI and 52\% of IPI suggest the model may miss comorbid or context-dependent syndromes, potentially affecting treatment accuracy. MMIR-TCM should therefore be positioned as an assistive system under professional supervision rather than an autonomous agent. Its clinical relevance lies in a human–machine collaborative paradigm that reduces documentation and retrieval burden and allows clinicians to focus on judgment and communication. Feedback from experienced TCM physicians supports this collaborative mode as a feasible and responsible path to implementation.

\section{Conclusion}
We present MMIR-TCM, an end-to-end multimodal framework that combines robust tongue region extraction, structured tongue diagnosis generation, and RAG for evidence-anchored prescription recommendations. Evaluated on the multi-center MedTCM corpus with the TDEU metric, the proposed pipeline improves attribute-level diagnostic fidelity and reduces unsupported recommendations by grounding prescriptions in deidentified clinical precedents and documented formula rationale, thereby enhancing factuality and clinician auditability. Limitations include reliance on retrospective, site biased records and sensitivity to image quality and metadata completeness; future work will focus on prospective validation, broader multi center data collection, integration of additional diagnostic modalities, and interface designs that present provenance and uncertainty to clinicians. Overall, MMIR-TCM offers a reproducible path toward multimodal clinical decision support in TCM practice.

\section{REFERENCES}
\bibliographystyle{IEEEtran}
\bibliography{ref}

\end{document}